\documentclass{article}
\usepackage{spconf,amsmath,graphicx}
\usepackage[margin=1in,bottom=1.10in,footskip=0.6in,columnsep=0.32in]{geometry} 
\usepackage{enumitem}
\usepackage{subcaption}    
\setlist{nosep, leftmargin=14pt}
\usepackage{mwe} 

\makeatletter
\renewcommand\section{\@startsection {section}{1}{\z@}%
  {2.5ex plus .8ex minus .4ex}
  {1.5ex plus .5ex minus .3ex}
  {\normalfont\Large\bfseries}}

\renewcommand\subsection{\@startsection {subsection}{2}{\z@}%
  {2.0ex plus .6ex minus .3ex}
  {1.0ex plus .3ex minus .2ex}
  {\normalfont\large\bfseries}}

\renewcommand\subsubsection{\@startsection {subsubsection}{3}{\z@}%
  {1.5ex plus .4ex minus .2ex}
  {0.8ex plus .2ex minus .1ex}
  {\normalfont\normalsize\bfseries}}
\makeatother

\title{Survival Modeling from Whole Slide Images via Patch-Level Graph Clustering and Mixture Density Experts}
\name{
\begin{tabular}{c}
    \textit{Ardhendu Sekhar}, \textit{Vasu Soni}, \textit{Keshav Aske}, \textit{Garima Jain}, \textit{Pranav Jeevan}, \textit{Amit Sethi}
\end{tabular}
}
\address{Indian Institute of Technology Bombay}
\begin{document}
%
\maketitle
\begin{abstract}
We present a modular framework for predicting cancer-specific survival from whole-slide pathology images (WSIs). The approach integrates four main components: (i) Quantile-Based  Patch Filtering, which employs quantile-based thresholding to identify prognostically informative tissue regions; (ii) Graph Regularized Patch Clustering using $k$-NN graph to model phenotype-level heterogeneity through spatial–morphological coherence; (iii) Hierarchical Feature Aggregation for learning intra- and inter-cluster dependencies; and (iv) an Expert-Guided Mixture Density Modeling module to estimate complex survival distributions using Gaussian distributions. The proposed model achieves a concordance index of $0.653{\pm}0.037$ on TCGA-LUAD, $0.719{\pm}0.011$ on TCGA-KIRC, and $0.733{\pm}0.037$ on TCGA-BRCA, surpassing current state-of-the-art methods.
\end{abstract}
\section{Introduction}
\label{sec:intro}
Precise survival prediction from histopathology images is essential for personalized oncology, enabling effective treatment planning and risk assessment. Whole-slide images (WSIs) encapsulate detailed tissue morphology—covering tumor structure and its microenvironmental context—but their enormous resolution and absence of localized labels pose challenges for direct modeling. Consequently, patch-based multiple instance learning (MIL) has emerged as the prevailing approach, segmenting each slide into smaller tiles for weakly supervised analysis.

Although MIL has proven effective for cancer grading and subtyping, survival prediction further requires modeling long-range dependencies and subtle morphological cues across distant tissue regions. Traditional methods like the Cox proportional hazards model fail to capture such complex, non-linear relationships, while previous deep learning models often lack interpretability and spatial awareness.


To overcome these challenges, we introduce a unified framework comprising four key modules:
(1) Quantile-Based Patch Filtering, which adaptively selects prognostically informative tissue regions via quantile thresholding;
(2) Graph-Regularized Patch Clustering, designed to capture spatial–morphological coherence among relevant patches;
(3) Hierarchical Feature Aggregation, enabling the learning of intra- and inter-cluster dependencies; and
(4) an Expert-Guided Mixture Density Modeling component for estimating complex survival distributions.
Collectively, these modules facilitate interpretable and accurate survival prediction by integrating spatial organization, phenotypic representation, and probabilistic outcome modeling.

\section{Related Work}

Weakly supervised learning plays a pivotal role in WSI-based survival prediction, primarily utilizing multiple instance learning (MIL). Early approaches such as ABMIL~\cite{ilse2018attentionbaseddeepmultipleinstance}, DSMIL~\cite{li2021dualstreammultipleinstancelearning}, and CLAM~\cite{lu2020dataefficientweaklysupervised} employed attention pooling to identify prognostic patches but neglected spatial dependencies. Graph-based models like PatchGCN~\cite{chen2021slideimages2dpoint} incorporated spatial relationships, while hierarchical transformer models such as HIPT~\cite{chen2022scalingvisiontransformersgigapixel}, HGT~\cite{10.1007/978-3-031-43987-2_72}, and TransMIL~\cite{shao2021transmiltransformerbasedcorrelated} captured multi-scale representations at significant computational cost. More recent multimodal frameworks—HiLa-X~\cite{CuiJia_HiLa_MICCAI2025}, PathoGen-X~\cite{10981028}, and CoC~\cite{ZhoHai_CoC_MICCAI2025}—combine histopathology with textual, genomics, or methylation data modalities, while SCMIL~\cite{Yang_2024} models survival outcomes using a Normal distribution. Nevertheless, these methods lack explicit mechanisms to disentangle latent phenotypes or leverage structural coherence. Our expert-guided mixture density module bridges this gap by coupling phenotype-aware clustering with expert-based survival modeling, where specialized experts capture distinct prognostic cues and a gating network adaptively integrates their outputs for enhanced interpretability.

\begin{figure*}[t]
    \centering
    \includegraphics[width=0.90\textwidth]{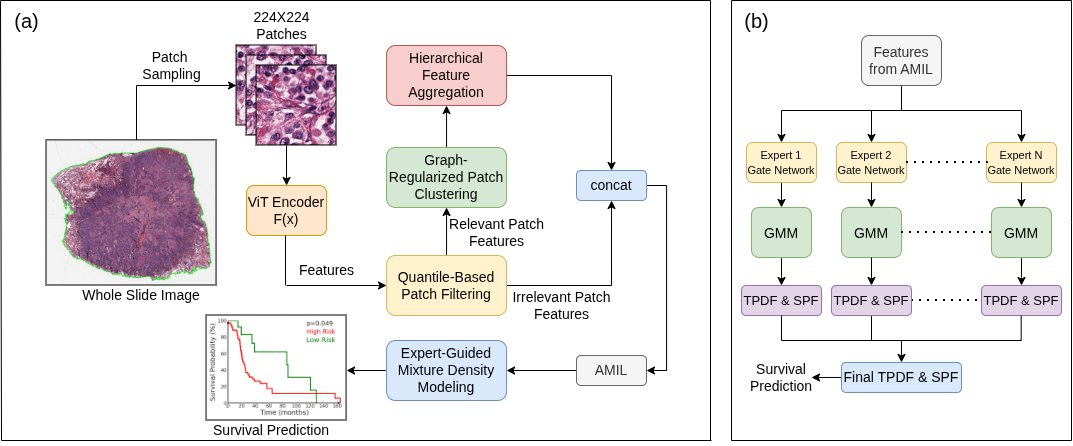}
    \caption{\textbf{(a)} Proposed survival model. \textbf{(b)} Expert-guided Mixture Density Modeling block.}
    \label{fig:pipeline}
    \label{fig:pipeline}
\end{figure*}

\section{Methodology}\label{sec:method}

Figure~\ref{fig:pipeline} presents an overview of the proposed model. Each Whole Slide Image (WSI) is partitioned into non-overlapping $224\times224$ patches at the highest magnification (40$\times$). After excluding background and non-tissue areas, the remaining tissue patches are embedded using a histopathology foundation model\cite{kang2023benchmarkingselfsupervisedlearningdiverse} i.e. a Vision Transformer (ViT) feature extractor $F(x)$, pretrained with large-scale self-supervised learning\cite{chen2021empiricalstudytrainingselfsupervised} on diverse WSI datasets, generates the corresponding patch feature matrix:
\vspace{-0.5em}
\begin{equation}
\mathbf{P}_{\text{feat}} \in \mathrm{R}^{n \times d}
\end{equation}
Here, $n$ represents the total number of extracted patches, and $d$ denotes the dimensionality of each feature vector.


The overall framework consists of four core modules: Quantile-Based Patch Filtering (QPF), Graph-Regularized Patch Clustering (GRPC), Hierarchical Feature Aggregation (HFA), and Expert-Guided Mixture Density Modeling (EGMDM).

\vspace{-0.5em}
\subsection{Quantile-Based Patch Filtering (QPF)}
\vspace{-0.3em}
To emphasize diagnostically significant regions, a learnable quantile-based Multi-Layer Perceptron (MLP) identifies task-relevant patches based on importance scores computed from model logits. For each WSI, a threshold $\tau_q$ corresponding to the $q$-quantile of logits is computed, preserving the top $(1-q)\times100\%$ of patches.
\vspace{-0.5em}
\begin{equation}
\mathcal{P}_{\text{sel}}=\{\mathbf{P}_i \mid \text{logit}_i>\tau_q\},\quad
\mathcal{P}_{\text{rem}}=\{\mathbf{P}_i \mid \text{logit}_i\le\tau_q\}.
\label{eq:sel}
\end{equation}
This dynamic filtering removes irrelevant patches, enabling subsequent modules to concentrate on semantically rich regions relevant to survival outcomes.

\vspace{-0.3em}
\subsection{Graph-Regularized Patch Clustering (GRPC)}
\vspace{-0.3em}
The selected patches $\mathcal{P}_{\text{sel}}\in\mathrm{R}^{m\times d}$ are clustered for both morphological and spatial consistency. Features and coordinates are normalized to a common scale, and cosine similarity measures morphological relationships:
\vspace{-0.5em}
\begin{equation}
S_{\text{morph}}(i,j)=
\frac{\langle \mathbf{P}_i, \mathbf{P}_j\rangle}{\|\mathbf{P}_i\|\,\|\mathbf{P}_j\|},
\label{eq:smorph}
\end{equation}
Meanwhile, spatial similarity is measured using an exponential kernel applied to the Euclidean distances $D_{ij}$ computed between the normalized patch coordinates:
\vspace{-0.5em}
\begin{equation}
S_{\text{spatial}}(i,j)=\exp\!\left(-\frac{D_{ij}}{\sigma_D}\right),
\quad \text{where } \sigma_D=\text{std}(D)+\varepsilon.
\label{eq:sspatial}
\end{equation}
\vspace{-1em}
A unified similarity matrix combines both cues:
\begin{equation}
S=\omega_{\text{morph}}S_{\text{morph}}+\omega_{\text{spatial}}S_{\text{spatial}}
\label{eq:sumS}
\end{equation}
Using the similarity matrix $S$, a $k$-nearest-neighbor (k-NN) graph is constructed, and GPU-accelerated K-Means guided by the graph topology clusters patches into $G$ groups ${L_1,\dots,L_G}$. Each cluster reflects a specific tissue microenvironment—like tumor epithelium, stroma, or necrosis—providing a biologically meaningful and spatially consistent structure within the slide.

\vspace{-0.3em}
\subsection{Hierarchical Feature Aggregation (HFA)}
\vspace{-0.3em}
To model dependencies within clusters, a two-level attention mechanism is employed.

\textbf{Intra-cluster attention.}
Each cluster $L_i$ is processed using multi-head self-attention (MHSA)~\cite{vaswani2017attention} to capture fine-grained local relationships among patches:
\vspace{-0.5em}
\begin{equation}
L'_i=\text{LayerNorm}\!\big(L_i+\text{MHSA}(L_i)\big).
\label{eq:intra}
\end{equation}
This step improves contextual understanding within morphologically similar regions.

\textbf{Inter-cluster attention.}
The refined clusters are condensed into representative embeddings.
\vspace{-0.5em}
\begin{equation}
R_i=\frac{1}{|L'_i|}\sum_{\mathbf{x}\in L'_i}\mathbf{x},
\label{eq:rep}
\end{equation}
which are subsequently passed through another MHSA layer to capture global relationships across regions:
\vspace{-0.5em}
\begin{equation}
R'=\text{LayerNorm}\!\big(R+\text{MHSA}(R)\big).
\label{eq:inter}
\end{equation}

Intra-cluster features are concatenated as \(\widetilde{P}=\mathrm{Concat}(L'_g)_{g=1}^{G}\). Averaging \(R'\) across clusters produces a broadcast descriptor \(R'_{\mathrm{exp}}\), expanded to match \(\widetilde{P}\) and added residually to yield \(\widehat{P}\). Finally, \(\widehat{P}\) is concatenated with unselected irrelavant patch features to form \(\mathcal{P}_{\text{final}}\).

 \textbf{WSI-level aggregation.} Slide-level representation is obtained through attention pooling as in AMIL~\cite{ilse2018attentionbaseddeepmultipleinstance}.
\vspace{-0.5em}
\begin{equation}
\begin{aligned}
\mathbf{z}_{\text{WSI}} &= \sum_i \alpha_i\,\mathcal{P}_{\text{final},i},\\
\alpha_i &= \text{softmax}\!\big(W_a\,\tanh(W_h\,\mathcal{P}_{\text{final},i}^{\top})\big),
\end{aligned}
\label{eq:wsiagg}
\end{equation}
where $\mathbf{z}_{\text{WSI}}$ represents the aggregated prognostic feature of the WSI.

\vspace{-0.3em}
\subsection{Expert-Guided Mixture Density Modeling\\(EGMDM)}
\vspace{-0.3em}
The Expert-Guided Mixture Density Modeling block (Figure~\ref{fig:pipeline}b) estimates continuous survival probabilities by modeling survival times, with each expert focusing on distinct WSI features $\mathbf{z}_{\text{WSI}}$. Following the mixture-of-experts paradigm~\cite{bishop2006pattern} and SurvMDN~\cite{han2022survivalmixturedensitynetworks}, it constructs Gaussian mixtures to represent diverse survival patterns.

The survival time $y$ corresponding to a whole slide image is modeled on its features $\mathbf{z}_{\text{WSI}}$, as illustrated in Figure~\ref{fig:pipeline}b. Each expert $e$ within the mixture outputs a Gaussian Mixture Model (GMM) over $y$:
\begin{equation}
p(y\mid \mathbf{z}_{\text{WSI}},e)=\sum_{k=1}^{K}
\lambda^{(e)}_k(\mathbf{z}_{\text{WSI}})\,
\mathcal{N}\!\big(y\mid \mu^{(e)}_k, \sigma^{(e)2}_k\big),
\label{eq:gmm}
\end{equation}
where each $k$ component follows a Gaussian distribution with mean \(\mu^{(e)}_k\) and standard deviation \(\sigma^{(e)}_k\), 
while mixture weights \(\lambda^{(e)}_k\) are computed via softmax over the \(K\) components. Learnable anchors (\((P_\mu, P_\sigma) \in \mathrm{R}^K\)) represent shared mean and standard deviation vectors, linearly transformed for each expert:
\vspace{-0.5em}
\begin{equation}
\mu^{(e)} = W^{(e)}_\mu P_\mu,\qquad
\sigma^{(e)} = \text{softplus}\!\big(W^{(e)}_\sigma P_\sigma\big),
\label{eq:anchors}
\end{equation}
The survival time $y$ is mapped to an unconstrained variable $t$ via a softplus transformation $t=\log(1+e^{y})$, which ensures non-negativity in its Gaussian domain and produces the Jacobian factor $\frac{dy}{dt}=\frac{e^{t}}{e^{t}-1}$. Finally, a gating network $G(\mathbf{z}_{\text{WSI}})$ assigns softmax probabilities across experts, yielding the overall survival probability functions:
\vspace{-0.5em}
\begin{align}
\text{TPDF}(t\mid \mathbf{z}_{\text{WSI}}) &= \frac{dy}{dt}
\sum_{e,k} G_e(\mathbf{z}_{\text{WSI}})\,\lambda^{(e)}_k\, \nonumber \\
&\quad \times \mathcal{N}\!\big(g^{-1}(t)\mid \mu^{(e)}_k,\sigma^{(e)2}_k\big)
\label{eq:tpdf}
\end{align}
\vspace{-1em}
\begin{equation}
\scalebox{0.95}{\mbox{$\displaystyle
\text{SPF}(t\mid \mathbf{z}_{\text{WSI}}) = 1 -
\sum_{e,k} G_e(\mathbf{z}_{\text{WSI}})\,\lambda^{(e)}_k\,
\Phi\!\left(\frac{g^{-1}(t)-\mu^{(e)}_k}{\sigma^{(e)}_k}\right)
$}}
\label{eq:spf}
\end{equation}
where $\Phi$ is derived by integrating $\mathcal{N}$, TPDF denotes the transformed probability density of death, and SPF represents the survival probability at time $t$.
\subsection{Training Objective and Regularization}
\vspace{-0.3em}
The model is trained using a negative \ log-likelihood loss ($\mathcal{L}_{\text{NLL}}$) that accounts for both censored ($c=0$ i.e. event not yet occurred) and uncensored ($c=1$, i.e. event occurred) cases. Given an observed time $t_d$ and censoring indicator $c \in {0,1}$,
\vspace{-0.5em}
\begin{equation}
\resizebox{0.95\linewidth}{!}{$
\mathcal{L}_{\text{NLL}}=
-\,c\,\log\!\big(\text{TPDF}(t_d\mid \mathbf{z}_{\text{WSI}})\big)
- (1-c)\,\log\!\big(\text{SPF}(t_d\mid \mathbf{z}_{\text{WSI}})\big)
$}
\label{eq:nll}
\end{equation}
To promote interpretability and specialization, a gating entropy loss ($\mathcal{L}_{\text{ent}}$) is introduced for diverse expert usage.
\vspace{-1.0em}
\begin{equation}
\mathcal{L}_{\text{ent}} = -{\textstyle\sum_{e}}\, G_e(\mathbf{z}_{\text{WSI}})\,
\log\!\big(G_e(\mathbf{z}_{\text{WSI}})\big)
\label{eq:lent}
\end{equation}
The total loss is:
\vspace{-0.5em}
\begin{equation}
\mathcal{L}_{\text{total}}=
\mathcal{L}_{\text{NLL}}+\lambda_{\text{ent}}\,\mathcal{L}_{\text{ent}}.
\label{eq:total}
\end{equation}
\vspace{-2.5em}

\begin{figure*}[t]
    \centering
    \includegraphics[width=0.90\textwidth]{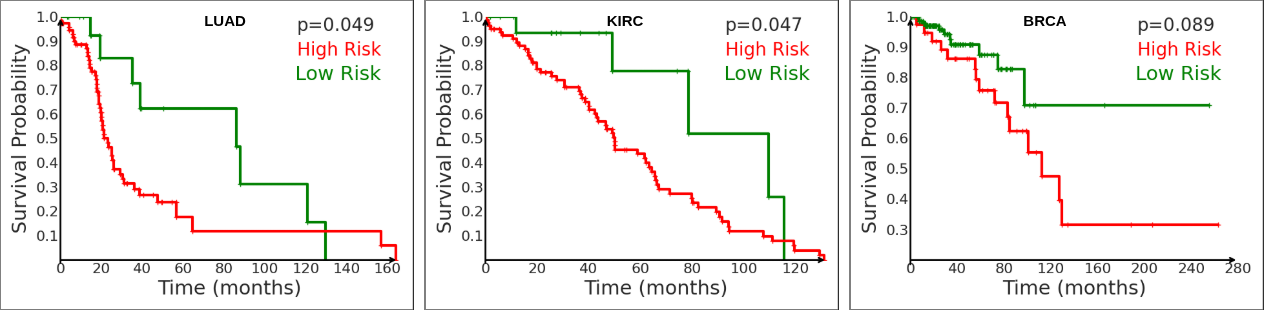}
    \caption{Kaplan–Meier survival plots illustrating distinct high and low-risk group separation across TCGA cohorts.}
    \label{fig:km_combined}
\end{figure*}

\begin{table*}[t]
\centering
\captionsetup{font=small}
\caption{Left: TCGA results (TDC, higher is better). Right: Ablations on $q$ in QPF, GRPC, HFA and $e$ in EGMDM. Best TDC values in \textbf{bold}.}
\label{tab:main+abl}
\begin{subtable}[t]{0.60\textwidth}
\centering
\scriptsize
\setlength{\tabcolsep}{5pt}
\renewcommand{\arraystretch}{1.05}
\caption{TCGA results. Modality:I=image,G=genomic,M=methylation,L=language prompt.}
\label{tab:results}
\begin{tabular}{l|c|c|c|c|c}
\hline
\textbf{Method} & \textbf{Modality} & \textbf{LUAD} & \textbf{KIRC} & \textbf{BRCA} & \textbf{MEAN} \\
\hline
AMIL~\cite{ilse2018attentionbaseddeepmultipleinstance} & I & $0.624 \pm 0.038$ & $0.638 \pm 0.065$ & $0.654 \pm 0.013$ & $0.639$ \\
CLAM~\cite{lu2020dataefficientweaklysupervised}        & I & $0.605 \pm 0.051$ & $0.676 \pm 0.032$ & $0.632 \pm 0.045$ & $0.638$ \\
DSMIL~\cite{li2021dualstreammultipleinstancelearning}  & I & $0.593 \pm 0.065$ & $0.655 \pm 0.031$ & $0.618 \pm 0.021$ & $0.622$ \\
TransMIL~\cite{shao2021transmiltransformerbasedcorrelated} & I & $0.545 \pm 0.037$ & $0.632 \pm 0.036$ & $0.591 \pm 0.013$ & $0.589$ \\
SetMIL~\cite{zhao2022setmil}              & I & $0.620 \pm 0.027$ & $0.694 \pm 0.052$ & $0.635 \pm 0.027$ & $0.650$ \\
PatchGCN~\cite{chen2021slideimages2dpoint}             & I & $0.587 \pm 0.019$ & $0.675 \pm 0.045$ & $0.582 \pm 0.037$ & $0.615$ \\
HIPT~\cite{chen2022scalingvisiontransformersgigapixel} & I & $0.549 \pm 0.025$ & $0.640 \pm 0.037$ & $0.625 \pm 0.046$ & $0.605$ \\
HGT~\cite{10.1007/978-3-031-43987-2_72}                & I & $0.607 \pm 0.058$ & $0.648 \pm 0.018$ & $0.638 \pm 0.022$ & $0.631$ \\
SCMIL~\cite{Yang_2024}                                  & I & $0.622 \pm 0.015$ & $0.688 \pm 0.037$ & $0.674 \pm 0.048$ & $0.661$ \\
\hline
PathoGen-X~\cite{10981028}                             & I+G & $0.620 \pm 0.008$ & $- - -$          & $0.670 \pm 0.020$ & $0.645$ \\
CoC~\cite{ZhoHai_CoC_MICCAI2025}                       & I+M & $- - -$          & $0.709 \pm 0.048$ & $0.654 \pm 0.036$ & $0.682$ \\
HiLa~\cite{CuiJia_HiLa_MICCAI2025}                     & I+L & $0.643 \pm 0.055$ & $- - -$          & $0.659 \pm 0.044$ & $0.651$ \\
\hline
\textbf{Ours}                                          & \textbf{I} & \textbf{0.653} $\pm$ \textbf{0.037} & \textbf{0.719} $\pm$ \textbf{0.011} & \textbf{0.733} $\pm$ \textbf{0.037} & \textbf{0.700} \\
\hline
\end{tabular}
\end{subtable}\hfill
\begin{subtable}[t]{0.37\textwidth}
\centering
\scriptsize
\setlength{\tabcolsep}{3.5pt}
\renewcommand{\arraystretch}{1.05}
\caption{Ablations.}
\label{tab:ablation}
\begin{tabular}{l|c|c|c}
\hline
\textbf{Variant} & \textbf{LUAD} & \textbf{KIRC} & \textbf{BRCA} \\
\hline
$q$=0.5 in QPF  & 0.634$\pm$0.022 & 0.700$\pm$0.052 & 0.720$\pm$0.046 \\
$q$=0.75 in QPF & 0.625$\pm$0.041 & 0.690$\pm$0.037 & 0.710$\pm$0.019 \\
w/o QPF   & 0.626$\pm$0.053 & 0.691$\pm$0.065 & 0.711$\pm$0.027 \\
w/o GRPC, HFA   & 0.621$\pm$0.046 & 0.685$\pm$0.050 & 0.705$\pm$0.011 \\
$e$=1 in EGMDM & 0.638$\pm$0.049 & 0.698$\pm$0.013 & 0.717$\pm$0.057 \\
\hline
Ours (All) & \textbf{0.653}$\pm$\textbf{0.037} & \textbf{0.719}$\pm$\textbf{0.011} & \textbf{0.733}$\pm$\textbf{0.037} \\
\hline
\end{tabular}
\end{subtable}
\end{table*}

\section{Experimental Setup}

\subsection{Datasets}

We assess the proposed model on three TCGA cohorts: Lung Adenocarcinoma (LUAD; 459 WSIs), Kidney Renal Clear Cell Carcinoma (KIRC; 509 WSIs), and Breast Invasive Carcinoma (BRCA; 956 WSIs)~\cite{tcga2013pan}. Each cohort comprises WSIs annotated with overall survival time, recorded in months from diagnosis to death (uncensored, c = 1) or until the last follow-up (censored, c = 0).
\vspace{-0.5em}
\subsection{Training Configuration}
The quantile of patch filtering module is fixed at $q$=0.25, selecting the top $75\%$ of patches by their logits. The k-NN graph built on similarity matrix $S$, connects each node to 10 neighbors, clustering patch indices into $G$ groups of 64 each and those corresponding patches are processed by an 8-head MHSA in the HFA module. The Expert-Guided Mixture Density block employs five experts, each modeling a \(K\!=\!100\) Gaussian mixture. Training runs for 20 epochs with Adam (lr \(2\times10^{-4}\), weight decay \(1\times10^{-3}\)), batch size 1. Performance is reported as mean~\(\pm\)~std of the Time-Dependent Concordance Index (TDC)~\cite{han2022survivalmixturedensitynetworks} over 5-fold cross-validation on an NVIDIA A6000 GPU.

\section{Results and Analysis}

\subsection{Quantitative Results}


Table~\ref{tab:results} presents the model’s performance across all datasets. Our framework consistently attains the highest mean TDC, reflecting superior discrimination and calibration. Histopathology-based multiple instance learning (MIL) models—AMIL~\cite{ilse2018attentionbaseddeepmultipleinstance}, CLAM~\cite{lu2020dataefficientweaklysupervised}, DSMIL~\cite{li2021dualstreammultipleinstancelearning}, TransMIL~\cite{shao2021transmiltransformerbasedcorrelated}, and SetMIL~\cite{zhao2022setmil}—served as the backbones for the expert-guided mixture density module, with results reported for comparison. Survival-oriented MIL approaches such as HIPT~\cite{chen2022scalingvisiontransformersgigapixel}, Patch-\hspace{0pt}GCN~\cite{10981028}, HGT~\cite{10.1007/978-3-031-43987-2_72}, and SCMIL~\cite{Yang_2024} act as benchmarks. As multimodal frameworks~\cite{10981028,ZhoHai_CoC_MICCAI2025,CuiJia_HiLa_MICCAI2025} lack open-source implementations, their reported results are drawn from the original publications. By combining quantile-based patch filtering, graph clustering, hierarchical aggregation, and expert-guided mixture density modeling—each expert capturing distinct morphological or contextual cues—our approach effectively identifies survival-relevant regions and achieves state-of-the-art results across all datasets, even surpassing multimodal counterparts while using only histopathology data.
\vspace{-0.5em}
\subsection{Ablation Studies and Interpretability}
Ablation experiments evaluated the effects of Quantile-Gated Patch Filtering, Graph-Regularized Patch Clustering, Hierarchical Feature Aggregation, and Expert-Guided Mixture Density Modeling (Table~\ref{tab:ablation}). A quantile of $q{=}0.25$ offered the best trade-off, retaining prognostic regions while suppressing noise. Removing any module reduced TDC, confirming their complementary roles in denoising and spatial reasoning. Single-expert models underperformed, whereas five experts improved results, highlighting the value of multi-expert modeling. For interpretability, patients were stratified into high- and low-risk groups using predicted survival scores. Kaplan–Meier curves (Fig.~\ref{fig:km_combined}) showed clear separation (log-rank $p{=}0.049$ LUAD, $p{=}0.047$ KIRC, $p{=}0.089$ BRCA), validating prognostic relevance and regional localization.

\section{Conclusion}
We introduced a unified framework for WSI-based survival prediction that combines quantile-regularized patch filtering, graph-based patch clustering, hierarchical feature aggregation, and expert-guided mixture density modeling. By capturing local–global tissue dependencies through clustered attention and modeling survival distributions using mixture experts, the approach enhances both calibration and discrimination. Experiments on TCGA-LUAD, TCGA-KIRC, and TCGA-BRCA demonstrate consistent improvements in time-dependent concordance compared to pathology-based and multimodal baselines. Future directions include extending the framework to multimodal and uncertainty-aware patient-level survival prediction for improved clinical robustness.

\section{Compliance with ethical standards}
\label{sec:ethics}

This study employed publicly available human data from~\cite{tcga2013pan}; thus, no separate ethical approval was required under the open-access license.

\section{Acknowledgments}
This study’s findings are based on data from the TCGA Research Network~\cite{tcga2013pan}.

\bibliographystyle{IEEEbib}
\bibliography{strings,refs}

@misc{kang2023benchmarkingselfsupervisedlearningdiverse,
      title={Benchmarking Self-Supervised Learning on Diverse Pathology Datasets}, 
      author={Mingu Kang and Heon Song and Seonwook Park and Donggeun Yoo and Sérgio Pereira},
      year={2023},
      eprint={2212.04690},
      archivePrefix={arXiv},
      primaryClass={cs.CV},
      url={https://arxiv.org/abs/2212.04690}, 
}

@misc{chen2021empiricalstudytrainingselfsupervised,
      title={An Empirical Study of Training Self-Supervised Vision Transformers}, 
      author={Xinlei Chen and Saining Xie and Kaiming He},
      year={2021},
      eprint={2104.02057},
      archivePrefix={arXiv},
      primaryClass={cs.CV},
      url={https://arxiv.org/abs/2104.02057}, 
}

@misc{ilse2018attentionbaseddeepmultipleinstance,
      title={Attention-based Deep Multiple Instance Learning}, 
      author={Maximilian Ilse and Jakub M. Tomczak and Max Welling},
      year={2018},
      eprint={1802.04712},
      archivePrefix={arXiv},
      primaryClass={cs.LG},
      url={https://arxiv.org/abs/1802.04712}, 
}

@misc{han2022survivalmixturedensitynetworks,
      title={Survival Mixture Density Networks}, 
      author={Xintian Han and Mark Goldstein and Rajesh Ranganath},
      year={2022},
      eprint={2208.10759},
      archivePrefix={arXiv},
      primaryClass={cs.LG},
      url={https://arxiv.org/abs/2208.10759}, 
}

@inbook{Yang_2024,
   title={SCMIL: Sparse Context-Aware Multiple Instance Learning for Predicting Cancer Survival Probability Distribution in Whole Slide Images},
   ISBN={9783031720833},
   ISSN={1611-3349},
   url={http://dx.doi.org/10.1007/978-3-031-72083-3_42},
   DOI={10.1007/978-3-031-72083-3_42},
   booktitle={Medical Image Computing and Computer Assisted Intervention – MICCAI 2024},
   publisher={Springer Nature Switzerland},
   author={Yang, Zekang and Liu, Hong and Wang, Xiangdong},
   year={2024},
   pages={448–458} }

@book{bishop2006pattern,
  title     = {Pattern Recognition and Machine Learning},
  author    = {Christopher M. Bishop},
  year      = {2006},
  publisher = {Springer},
  address   = {New York},
  isbn      = {978-0-387-31073-2}
}

@misc{lu2020dataefficientweaklysupervised,
      title={Data Efficient and Weakly Supervised Computational Pathology on Whole Slide Images}, 
      author={Ming Y. Lu and Drew F. K. Williamson and Tiffany Y. Chen and Richard J. Chen and Matteo Barbieri and Faisal Mahmood},
      year={2020},
      eprint={2004.09666},
      archivePrefix={arXiv},
      primaryClass={eess.IV},
      url={https://arxiv.org/abs/2004.09666}, 
}

@misc{li2021dualstreammultipleinstancelearning,
      title={Dual-stream Multiple Instance Learning Network for Whole Slide Image Classification with Self-supervised Contrastive Learning}, 
      author={Bin Li and Yin Li and Kevin W. Eliceiri},
      year={2021},
      eprint={2011.08939},
      archivePrefix={arXiv},
      primaryClass={cs.CV},
      url={https://arxiv.org/abs/2011.08939}, 
}

@misc{chen2021slideimages2dpoint,
      title={Whole Slide Images are 2D Point Clouds: Context-Aware Survival Prediction using Patch-based Graph Convolutional Networks}, 
      author={Richard J. Chen and Ming Y. Lu and Muhammad Shaban and Chengkuan Chen and Tiffany Y. Chen and Drew F. K. Williamson and Faisal Mahmood},
      year={2021},
      eprint={2107.13048},
      archivePrefix={arXiv},
      primaryClass={eess.IV},
      url={https://arxiv.org/abs/2107.13048}, 
}

@misc{shao2021transmiltransformerbasedcorrelated,
      title={TransMIL: Transformer based Correlated Multiple Instance Learning for Whole Slide Image Classification}, 
      author={Zhuchen Shao and Hao Bian and Yang Chen and Yifeng Wang and Jian Zhang and Xiangyang Ji and Yongbing Zhang},
      year={2021},
      eprint={2106.00908},
      archivePrefix={arXiv},
      primaryClass={cs.CV},
      url={https://arxiv.org/abs/2106.00908}, 
}

@misc{chen2022scalingvisiontransformersgigapixel,
      title={Scaling Vision Transformers to Gigapixel Images via Hierarchical Self-Supervised Learning}, 
      author={Richard J. Chen and Chengkuan Chen and Yicong Li and Tiffany Y. Chen and Andrew D. Trister and Rahul G. Krishnan and Faisal Mahmood},
      year={2022},
      eprint={2206.02647},
      archivePrefix={arXiv},
      primaryClass={cs.CV},
      url={https://arxiv.org/abs/2206.02647}, 
}

@InProceedings{10.1007/978-3-031-43987-2_72,
    title="Multi-scope Analysis Driven Hierarchical Graph Transformer for Whole Slide Image Based Cancer Survival Prediction",
    booktitle="Medical Image Computing and Computer Assisted Intervention -- MICCAI 2023",
    year="2023",
    publisher="Springer Nature Switzerland",
    address="Cham",
    pages="745--754",
    isbn="978-3-031-43987-2"
}

@article{tcga2013pan,
  author    = {{TCGA Research Network}},
  title     = {The Cancer Genome Atlas Pan-Cancer analysis project},
  journal   = {Nature Genetics},
  volume    = {45},
  number    = {10},
  pages     = {1113--1120},
  year      = {2013},
  publisher = {Nature Publishing Group},
  doi       = {10.1038/ng.2764}
}

@INPROCEEDINGS{10981028,
  author={Krishna, Akhila and Kurian, Nikhil Cherian and Patil, Abhijeet and Parulekar, Amruta and P, Pranav Jeevan and Sethi, Amit},
  booktitle={2025 IEEE 22nd International Symposium on Biomedical Imaging (ISBI)}, 
  title={Pathogen-X: A Cross-Modal Genomic Feature Trans-Align Network for Enhanced Survival Prediction from Histopathology Images}, 
  year={2025},
  volume={},
  number={},
  pages={1-4},
  doi={10.1109/ISBI60581.2025.10981028}}

@InProceedings{ ZhoHai_CoC_MICCAI2025,
                 author = { Zhou, Haipeng AND Yang, Sicheng AND Yang, Sihan AND Qin, Jing AND Chen, Lei AND Zhu, Lei },
                 title = { { CoC: Chain-of-Cancer based on Cross-Modal Autoregressive Traction for Survival Prediction } }, 
                 booktitle = {Medical Image Computing and Computer Assisted Intervention -- MICCAI 2025},
                 year = {2025},
                 publisher = {Springer Nature Switzerland},
                 volume = { LNCS 15974 },
                 month = {October},
                 pages = { 85 -- 94 },
              }

@InProceedings{ CuiJia_HiLa_MICCAI2025,
                 author = { Cui, Jiaqi AND Wen, Lu AND Fei, Yuchen AND Liu, Bo AND Zhou, Luping AND Shen, Dinggang AND Wang, Yan },
                 title = { { HiLa: Hierarchical Vision-Language Collaboration for Cancer Survival Prediction } }, 
                 booktitle = {Medical Image Computing and Computer Assisted Intervention -- MICCAI 2025},
                 year = {2025},
                 publisher = {Springer Nature Switzerland},
                 volume = { LNCS 15964 },
                 month = {October},
                 pages = { 240 -- 250 },
              }

@inproceedings{zhao2022setmil,
  title     = {SETMIL: Spatial Encoding Transformer-Based Multiple Instance Learning for Pathological Image Analysis},
  author    = {Zhao, Yu and Lin, Zhenyu and Sun, Kai and Zhang, Yidan and Huang, Junzhou and Wang, Liansheng and Yao, Jianhua},
  booktitle = {Proc. MICCAI},
  year      = {2022},
  pages     = {66--76}
}

@inproceedings{vaswani2017attention,
  title = {Attention is All You Need},
  author = {Vaswani, A. et al.},
  booktitle = {Advances in Neural Information Processing Systems},
  year = {2017}
}

\end{document}